ADAPTIVE MULTIPLE OPTIMAL LEARNING FACTORS

FOR NEURAL NETWORK TRAINING

by

JESHWANTH CHALLAGUNDLA

Presented to the Faculty of the Graduate School of

The University of Texas at Arlington in Partial Fulfillment

of the Requirements

for the Degree of

MASTER OF SCIENCE IN ELECTRICAL ENGINEERING

THE UNIVERSITY OF TEXAS AT ARLINGTON

May 2015







Acknowledgements

Foremost, I would like to express my sincere gratitude to my advising professor Dr. Michael Manry for his support and the efforts he had put in to guide me in my research. I would also thank him for the time he had spent for me and the knowledge he provided me from his courses and lab meetings

I would thank Rohit Rawat for his valuable inputs and suggestions. I would also thank Gautam Eapi, Kanishka Tyagi, Son Nguyen for their support.

I would also thank my thesis committee members Dr. Ionnis D. Schizas and Dr. W. Alan Davis for reviewing my work.

I would like to thank and dedicate this thesis to my parents, Mr. Suresh and Mrs. Chandrakala and my brother, Prudhvi.

April 10, 2015



Abstract

ADAPTIVE MULTIPLE OPTIMAL LEARNING FACTORS ALGORITHM

FOR FEEDFORWARD NETWORKS

Jeshwanth Challagundla, M.S.

The University of Texas at Arlington, 2015

Supervising Professor: Michael Manry

There is always an ambiguity in deciding the number of learning factors that is really required for training a Multi-Layer Perceptron. This thesis solves this problem by introducing a new method of adaptively changing the number of learning factors computed based on the error change created per multiply. A new method is introduced for computing learning factors for weights grouped based on the curvature of the objective function. A method for linearly compressing large ill-conditioned Newton's Hessian matrices to smaller well-conditioned ones is shown. This thesis also shows that the proposed training algorithm adapts itself between two other algorithms in order to produce a better error decrease per multiply. The performance of the proposed algorithm is shown to be better than OWO-MOLF and Levenberg Marquardt for most of the data sets.



Table of Contents







List of Illustrations





List of Tables





Chapter 1

Introduction

1.1 Artificial Neural Networks

An artificial neural network (ANN) is a computational model inspired by biological

neurons. There are a number of different types of networks, but they are all characterized

by a set of nodes and connections between the nodes. These nodes are basic

computational units called neurons. The structure of an artificial neuron is shown in figure

1.1.

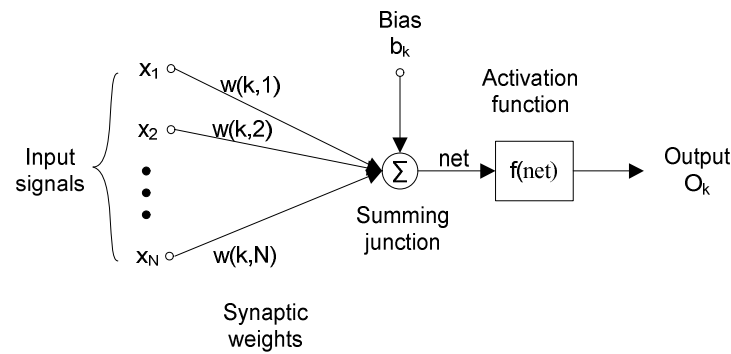

Figure 1.1 Nonlinear model of a neuron

The connections between nodes are called *synapses*. Every synapse has a

value associated with it called a synaptic weight. This weight is multiplied by the input of

a synapse in order to obtain the output at the other end of a synapse. A neuron also

consists of a summing junction that sums up outputs of all the synapses connected to it.

This is called the *net* value of the neuron. The net value is passed through an activation

function in order to obtain the output of a neuron. Typically used activation functions are

the piecewise linear function and nonlinear functions like sigmoid and hyperbolic tangent

[7], [1].



ANNs with their remarkable ability to derive meaning from complicated and imprecise data can be used to extract patterns and detect trends that are too complex to be noticed by either human or other computer techniques.

There are different kinds of ANNs based on the structural arrangement of neurons and the way data flows between them. All though there can be any number of arbitrary network configuration a few prominent ones are mentioned here. Feedforward networks were the first and simplest type of ANNs devised. In these networks, information moves only in one direction, forward, from the input nodes, through the hidden nodes and to the output nodes. There are no cycles or loops in these networks. A feedforward network with one hidden layer is called a single layer feedforward network, one with more than one hidden layer is called a multilayer feedforward network. ANNs with a feedback loop are called recurrent neural networks (NNs). ANNs which use both present as well as previous inputs are called time delayed NNs. ANNs which use radial basis functions which has built into a distance criterion with respect to a center are called radial basis function (RBF) networks.

Of all these different NN architectures, multilayer feedforward NNs are extensively used for function approximation and pattern recognition task's because of their special properties like universal approximation, approximation of Bayes discriminants, etc. The most commonly used multilayer feedforward network is the multilayer perceptron (MLP).

<div align="center">1.2 Properties of MLP</div>

*Universal Approximation:*

During 1970's the apparent ability of sufficiently elaborate feed-forward networks to approximate quite well nearly any function encountered in applications led investigators to wonder about the ultimate capabilities of such networks. Are the



successes observed reflective of some deep and fundamental approximation capability, or are they merely flukes, resulting from selective reporting and a fortuitous choice of problems? Are multilayer feedforward networks in fact inherently limited to approximating only some fairly special class of functions [58]?

However in 1988 Gallant and White [56] showed that a particular single hidden layer feed forward network using the monotone "cosine activation" is capable of yielding Fourier series approximation to a given function to any degree of accuracy. In 1989 Hect-Nielsen [57] proved the same for single hidden layer feedforward networks with logistic activation.

In 1989 Hornik, Stincombe and White [58] proved that standard multilayer feedforward networks are capable of approximating any measurable function to any desired degree of accuracy, in a very specific and satisfying sense. This is called the universal approximation theorem. It also implies that any lack of success in applications must arise from inadequate learning, insufficient number of hidden units or the lack of a deterministic relationship between input and target [59] [60]. However, the number of hidden units that are needed in the single hidden layer is still uncertain [43] [6].

*Approximating Bayes discriminants:*

Many publications [61] [62] [63] [64] [65] have shown that multilayer perceptron classifiers and conventional nonparametric Bayesian classifiers yield the same classification accuracy statistically. All these were empirical and hence are dependent on the data sets used.

It was proved theoretically in 1990 [66] that a multilayer perceptron trained using back propagation approximates the Bayes optimal discriminant functions for both two-class and multiclass recognition problems. Most importantly, it has been shown that the outputs of the multilayer perceptron approximate the a posteriori probability functions



when trained using back propagation for multiclass problems. This does not depend on the architecture of the network and is hence, applicable to any network that minimizes the mean squared-error measure.

Therefore a multilayer perceptron (MLP) can approximate both a Bayes classifier and a Bayes approximation [67] [68]. It must be noted that the accuracy of the MLP in approximating Bayes posterior probabilities improves as the number of training patterns increases.

## 1.3 Applications of MLP

The MLP is very good at fitting nonlinear functions and recognizing patterns [7] [1]; consequently they are used extensively in function approximation and pattern recognition.

*Pattern Recognition*

Pattern recognition is formally defined as the process whereby a received pattern/signal is assigned to one of a prescribed number of classes. A neural network performs pattern recognition by first undergoing a training session during which the network is repeatedly presented with a set of input patterns along with the category to which each particular pattern belongs. Later, the network is presented with a new pattern that has not been seen before, but which belongs to the same population of patterns used to train the network. The network is able to identify the class of that particular pattern because of the information that it had extracted from training data [6].

The MLP is used in pattern recognition problems such as speech recognition [14], character recognition [15], fingerprint recognition [16], face detection [17], classification and diagnostic prediction of cancer [52], handwritten zip code recognition [53], etc.



*Function Approximation*

Consider an unknown nonlinear input-output mapping described by a function f(·). In order to make up for lack of knowledge about the function f(·), a labeled data set is given. When the neural network is trained on this data set it tries to form an input-output mapping F(·) which is close enough to f(·) in a Euclidean sense over all the inputs in the labeled data set.

The MLP is used in approximation problems [7] like stock market forecasting [8], aviation prognostics [9], data mining [10, 11], filtering [12], control applications [13], energy systems [44], atmospheric science [45], hydrology [46], renewable energy systems [47], ecological modeling [48], electric load forecasting [49], rainfall-runoff modeling [50], weather forecasting [51] etc.

1.4 Problems with the MLP

In spite of the many applications of the MLP and its other advantages, it still has many problems that remain to be solved. A few of these problems are discussed here.

1. Training of neural networks is very sensitive to the initial values of weights. This problem can be partly solved by using net control. Net control is nothing but initializing the weights in such a way that the mean of net values is a small non zero number like 0.5 and their variance is a small value [54] such as 1.0.

2. First order algorithms lack affine invariance, i.e., two equivalent networks with different weight matrices that are trained using first order algorithms lose equivalence at the first iteration.

3. Second order algorithms use multiplies inefficiently and do not scale well. This occurs because second order algorithms calculate Hessian matrices which are



computationally expensive. As the network size increases the number of multiplies required grows exponentially.

4. There is always uncertainty in deciding the number of hidden layers and number of units in each hidden layer to be used [55].

5. Validation and confidence assessment is a problem [55]

## 1.5 Organization of this Thesis

Problems 2 and 3 can be partly solved by using Newton's algorithm to find small sets of network parameters in each iteration [3] [5] [70] [71].The objective of this thesis is to continue this work by designing an algorithm that can adapt itself in every iteration in such a way that the size of hessian computed changes based on the error decrease per multiply. To a great extent this reduces the computational burden of second order methods.

Chapter 2 reviews NN training including first order and second order algorithms and optimal learning factors. Chapter 3 discusses Output Weight Optimization, Newton's method for input weights and the OWO-MOLF algorithms in detail. In chapter 4 we present the motivation for this thesis and some lemmas. Chapter 5 introduces the Adaptive Multiple Optimal Learning Factors algorithm in detail and describes its advantages. Experimental results are included in chapter 6 and the thesis is concluded in chapter 7.



Chapter 2

Review of Neural Network Training

2.1 MLP Notation and Processing

The MLP is the most widely used type of neural network [32], Training a MLP involves solving a non-convex optimization problem to calculate the network weights. The single hidden layer MLP is the focus of this work because any continuous function can be approximated to arbitrary precision [33]. The architecture of a fully connected feed forward neural network is shown in figure 2.1.

Input weights $w(k,n)$ connects the $n^{th}$ input to the $k^{th}$ hidden unit. Output weights $w_{oh}(i,k)$ connect the $k^{th}$ hidden unit activation $O_p(k)$ to the $i^{th}$ output $y_p(m)$ with linear activation. Bypass weight $w_{oi}(i,n)$ connects the $i^{th}$ output to the $n^{th}$ input signal. The training data set $\{\mathbf{x_p}, \mathbf{t_p}\}$ consists of input vectors $\mathbf{x_p}$ that are initially of size N. One more input $x_p(N+1)$ whose value is always equal to 1 is added to handle the threshold for both input and hidden layers, so $\mathbf{x_p}=[x_p(1),x_p(2),....,x_p(N+1)]^T$. The desired output vector $\mathbf{t_p}$ contains M elements. 'p' is the pattern number that varies from 1 to $N_v$, where $N_v$ is the number of training vectors present in the data set. $N_h$ is the total number of hidden units present in the MLP hidden layer. The dimensions of the weight matrices $\mathbf{W}$, $\mathbf{W_{oh}}$ and $\mathbf{W_{oi}}$ are $N_h$ by (N+1), M by $N_h$ and M by (N+1) respectively [5].

The hidden layer net function vector, $\mathbf{n_p}$ and the actual output of the network, $\mathbf{y_p}$ can be written as [5],

$$\mathbf{n_p} = \mathbf{W} \cdot \mathbf{x_p} \qquad\qquad (2.1)$$

$$\mathbf{y_p} = \mathbf{W_{oi}} \cdot \mathbf{x_p} + \mathbf{W_{oh}} \cdot \mathbf{O_p} \qquad\qquad (2.2)$$



where the k[th] element of the hidden unit activation vector **O$_p$** is calculated as

$O_p(k)=f(n_p(k))$ and $_f(.)$ denotes the hidden layer activation function. Training an MLP

typically involves minimizing the mean squared error between the desired and the actual

network output, defined as

$$E = \frac{1}{N_v} \sum_{p=1}^{N_v} \sum_{i=1}^{M} \left[ t_p(i) - y_p(i) \right]^2 = \frac{1}{N_v} \sum_{p=1}^{N_v} E_p \tag{2.3}$$

$$E_p = \sum_{i=1}^{M} \left[ t_p(i) - y_p(i) \right]^2 \tag{2.4}$$

where, $E_p$ is the cumulative squared error for pattern p [5].

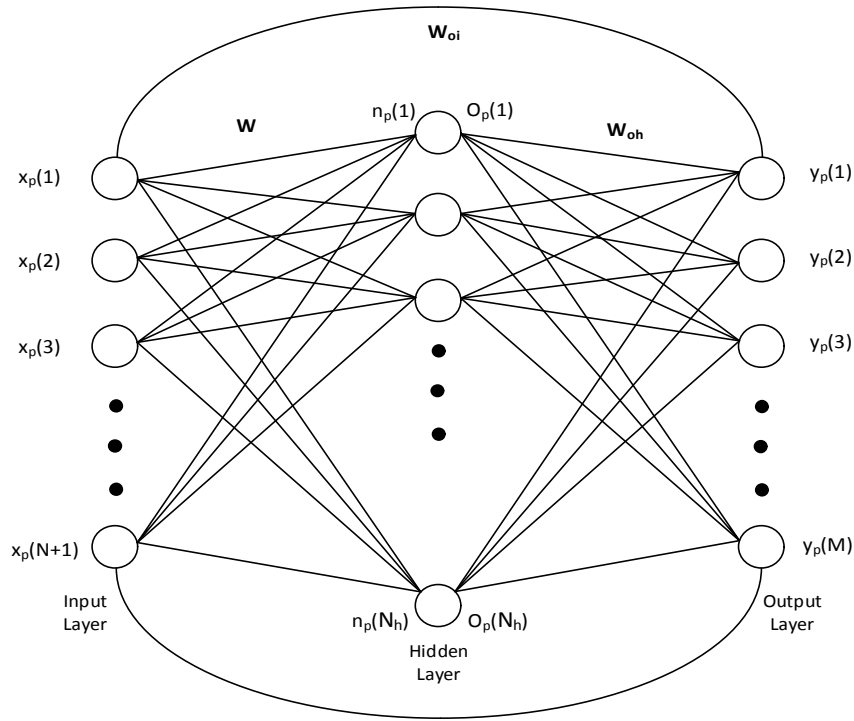

Figure 2.1 Fully connected MLP Structure



2.2 Review of First Order Training Algorithms

Training of neural networks consists of changing the weights and biases in order to make the computed output as close as possible to the desired output. It mainly involves the following two independent steps. First a search direction has to be determined. i.e., in what direction do we want to search in weight space for a new current point? Once the search direction has been found we have to decide how far to go in the specified search direction, i.e., a step size has to be determined.

Most of the optimization methods used to minimize error functions are based on the same strategy. The minimization is a local iterative process in which an approximation to the error function in a neighborhood of the current point in weight space is minimized. Neural network training depends mainly on the error function used. In this thesis mean squared error (2.3), (2.4) is used as the error function. Once the neural network is trained it will be able to make predictions on new data set whose outputs are unknown.

*2.2.1 Back Propagation*

Back propagation is a well-established method for calculating first order gradient matrices defined as

$$\mathbf{G} = \frac{-\partial \mathrm{E}}{\partial \mathbf{W}}$$

(2.5)

$$\mathbf{G}_{\mathbf{oh}} = \frac{-\partial \mathrm{E}}{\partial \mathbf{W}_{\mathbf{oh}}}$$

(2.6)

$$\mathbf{G}_{\mathbf{oi}} = \frac{-\partial \mathrm{E}}{\partial \mathbf{W}_{\mathbf{oi}}}$$

(2.7)

The gradients are calculated using the chain rule, in which delta functions are calculated for output and hidden units. Delta functions are derivatives of an error function



of a particular training pattern with respect to the net function. The delta functions of output and hidden units [73] are calculated using the equation (2.8) and (2.9)

$$\delta_{po}(i) = \frac{\partial E_p}{\partial y_p(i)} = 2(t_p(i) - y_p(i)) \tag{2.8}$$

$$\delta_p(k) = \frac{\partial E_p}{\partial n_p(k)} = f'(n_p(k)) \sum_{i=1}^{M} \delta_{po}(i) w_{oh}(i,k) \tag{2.9}$$

The elements of negative gradient matrices $\mathbf{G}$, $\mathbf{G_{oi}}$, $\mathbf{G_{oh}}$ of input, output and bypass weights $\mathbf{W}$, $\mathbf{W_{oh}}$, $\mathbf{W_{oi}}$ respectively are calculated from equation (2.10), (2.11) and (2.12)

$$g(k,n) = \frac{\partial E}{\partial w(k,n)} = \frac{1}{N_v} \sum_{p=1}^{N_v} \delta_p(k) x_p(n) \;\; ; \;\; \mathbf{G} = \frac{1}{N_v} \sum_{p=1}^{N_v} \boldsymbol{\delta_p} \, \mathbf{x_p}^\top \tag{2.10}$$

$$g_{oh}(i,k) = \frac{\partial E}{\partial w_{oh}(i,k)} = \frac{1}{N_v} \sum_{p=1}^{N_v} \delta_{po}(i) O_p(k) \;\; ; \;\; \mathbf{G_{oh}} = \frac{1}{N_v} \sum_{p=1}^{N_v} \boldsymbol{\delta_{po}} \mathbf{O_p}^\top \tag{2.11}$$

$$g_{oi}(i,n) = \frac{\partial E}{\partial w_{oi}(i,n)} = \frac{1}{N_v} \sum_{p=1}^{N_v} \delta_{po}(i) x_p(n) \;\; ; \;\; \mathbf{G_{oi}} = \frac{1}{N_v} \sum_{p=1}^{N_v} \boldsymbol{\delta_{po}} \, \mathbf{x_p}^\top \tag{2.12}$$

*Steepest Descent*

In steepest descent the weights are updated in the direction of negative gradients. The weight changes of input, output and bypass weights are calculated using the gradients from eq. (2.10), (2.11) and (2.12) and a *learning factor* z. The weights are updated as follows,

$$\mathbf{W} \leftarrow \mathbf{W} + z \cdot \mathbf{G} \tag{2.13}$$

$$\mathbf{W_{oh}} \leftarrow \mathbf{W_{oh}} + z \cdot \mathbf{G_{oh}} \tag{2.14}$$

$$\mathbf{W_{oi}} \leftarrow \mathbf{W_{oi}} + z \cdot \mathbf{G_{oi}} \tag{2.15}$$



*Optimal Learning Factor*

Learning factor z decides the rate of convergence of neural network training. Usually a small positive value for z will work, but convergence is likely to be slow. If z is too large the error E can increase [7]. In order to avoid this uncertainty a lot of heuristic scaling approaches have been introduced to modify the learning factors between iterations and thus speed up the rate of convergence. However using a Taylor's series for the error E, a non-heuristic Optimal Learning Factor (OLF) for OWO-BP (where OWO is Output Weight Optimization) can be calculated as,

$$z = \frac{-\partial E / \partial z}{\partial^2 E / \partial z^2} \tag{2.16}$$

where the numerator and denominator derivatives are evaluated at z=0. Assume that the learning factor z is used to update only the input weights **W**, as

$$\mathbf{W} \leftarrow \mathbf{W} + z \cdot \mathbf{G} \tag{2.17}$$

The expression for the second derivative of the error function with respect to the OLF is found using eq. (2.18), (2.19) as, [5]

$$\frac{\partial^2 E}{\partial z^2} = \sum_{k=1}^{N_h} \sum_{j=1}^{N_h} \sum_{n=1}^{N+1} g(k,n) \sum_{m=1}^{N+1} \frac{\partial^2 E}{\partial w(k,n) \partial w(j,m)} g(j,m) \tag{2.18}$$

$$= \sum_{k=1}^{N_h} \sum_{j=1}^{N_h} \mathbf{g}_\mathbf{k}^{\mathrm{T}} \mathbf{H}^{\mathbf{k,j}} \mathbf{g}_\mathbf{j} \tag{2.19}$$

where column vector $\mathbf{g_k}$ contains elements $g(k,n)$ of **G**, for all values of n. **H** is the reduced size input weight Hessian with $N_{iw}$ rows and columns, where $N_{iw}=(N+1)\cdot N_h$ is the total number of input weights. $\mathbf{H^{k,j}}$ contains elements of **H** for all input weights connected to the $j^{th}$ and $k^{th}$ hidden units and has size (N+1) by (N+1). When Gauss-Newton updates are used, elements of **H** are computed as



$$\frac{\partial^2 E}{\partial w(j,m) \partial w(k,n)} = \frac{2}{N_v} \sum_{p=1}^{N_v} \sum_{i=1}^{M} \frac{\partial y_p(i)}{\partial w(j,m)} \frac{\partial y_p(i)}{\partial w(k,n)} \qquad (2.20)$$

Eq. (2.16) and (2.20) show that (i) the OLF can be obtained from elements of the Hessian **H**; (ii) **H** contains useful information even when it is singular; and (iii) a smaller non-singular Hessian, $\partial^2 E / \partial z^2$ can be constructed using **H**. Therefore it may be advantageous to construct and use Hessians of intermediate size [5].

*OWO-BP Algorithm*

This is a two stage algorithm; in the first stage the Output Weight Optimization (OWO) technique is used in order to find out the output and bypass weights **W_oh** and **W_oi** respectively. In the second stage back propagation is used in order to compute the weight change matrix **ΔW** [72]. This weight change matrix is used to update the input weights **W**.

OWO is a technique used to find out the output weights [74]. Since the activations in the outputs are linear, OWO is equivalent to solving a set of linear equations. The actual outputs can be calculated from the below modified equation

$$\mathbf{y_p} = \mathbf{W_o} \cdot \mathbf{X_p} \qquad (2.21)$$

where the basis vector **X_p** is the augmented input of size $N_u$ formed by combining actual input and hidden unit activations $[\mathbf{x_p}, \mathbf{O_p}]^T$, where $N_u$ is equal to $N+N_h+1$. **W_o** includes both output and bypass weights $[\mathbf{W_{oi}} : \mathbf{W_{oh}}]$ and has a size of M by $N_u$. The output weights can be solved by equating $\partial E / \partial \mathbf{W_o}$ to zero; this is equivalent to solving the set of linear equations given in equation

$$\mathbf{C} = \mathbf{R} \cdot \mathbf{W_o}^T \qquad (2.22)$$

**R** is the autocorrelation matrix of the augmented input **X_p**, and **C** is the cross correlation of augmented input and desired output **t_p**.



$$C = \frac{1}{N_v} \sum_{p=1}^{N_v} \mathbf{X_p} \mathbf{t_p}^T \qquad (2.23)$$

$$R = \frac{1}{N_v} \sum_{p=1}^{N_v} \mathbf{X_p} \mathbf{X_p}^T \qquad (2.24)$$

Eq. (2.22) is most easily solved using orthogonal least squares (OLS) [2] [5].

Once the output weights are computed, back propagation is used to compute the negative gradient $\mathbf{G}$ of input weights as shown in eq. (2.5) (2.10) in order to calculate the input weight change matrix $\mathbf{\Delta W}$,

$$\mathbf{\Delta W} = z \cdot \mathbf{G} \qquad (2.25)$$

$$\mathbf{W} \leftarrow \mathbf{W} + z \cdot \mathbf{G} \qquad (2.26)$$

A description of the OWO-BP algorithm is given below. For every training epoch [5]

1. Solve the system of linear equations in (2.22) using OLS and update the output weights, $\mathbf{W_o}$

2. Find the negative Jacobian matrix $\mathbf{G}$ described in eq. (2.5) and (2.10)

3. Update the input weight matrix $\mathbf{W}$, using eq. (2.26)

This method is attractive for following reasons [5],

1. The training is faster than steepest descent and consumes fewer multiplies, since training weights connected to the outputs is equivalent to solving linear equations.

2. It helps us avoid some local minima

3. This method shows better performance when compared to the BP algorithm discussed before.



*Conjugate Gradient*

Conjugate Gradient is a first order training method that can be viewed as being intermediate between steepest descent and Newton's method. The conjugate gradient method has the following properties;

1. It minimizes quadratic error functions of *n* variables in *n* steps.

2. The conjugate gradient algorithm requires no Hessian matrix evaluation.

3. No matrix inversion or storage of an *n×n* matrix is required.

In the conjugate gradient method, weights **w**, are not directly updated using the gradient vector $\mathbf{g_k}$; instead they are updated using a direction vector $\boldsymbol{p_k}$ for the k$^{th}$ iteration, where,

$$\mathbf{g_k} = vec\left(\mathbf{G}, \mathbf{G_{oh}}, \mathbf{G_{oi}}\right) \tag{2.27}$$

$$\mathbf{w} = vec\left(\mathbf{W}, \mathbf{W_{oh}}, \mathbf{W_{oi}}\right) \tag{2.28}$$

$$\boldsymbol{p_k} = vec\left(\mathbf{P}, \mathbf{P_{oh}}, \mathbf{P_{oi}}\right) \tag{2.29}$$

The *vec* operator converts all the parameter matrices into a single vector. The subscript k indicates the iteration number. The direction vector is updated as follows,

$$\boldsymbol{p_{k+1}} = \mathbf{g_k} + \mathrm{B_1} \cdot \boldsymbol{p_k} \tag{2.30}$$

using a factor $\mathrm{B_1}$ calculated as [69],

$$\mathrm{B_1} = \frac{\mathbf{g_{k+1}}^{\mathrm{T}} \mathbf{g_{k+1}}}{\mathbf{g_k}^{\mathrm{T}} \mathbf{g_k}} \tag{2.31}$$

This direction vector, in turn, updates the weights as

$$\mathbf{w_{k+1}} = \mathbf{w_k} + \mathrm{z} \cdot \boldsymbol{p_k} \tag{2.32}$$



Since the weights are updated using the direction vector as opposed to the gradients themselves, the direction of descent is superior to that seen in steepest descent and converges in fewer iterations [34] [68] [69].

### 2.3 Review of Second Order Training Algorithms

Second order training of a MLP involves quadratic modeling of the error function. Second order training methods are preferred because of their fast convergence. They lead to problems like memory limitation, since the Hessian and Gradient matrices should be computed and stored. They also are computationally very expensive.

*Newton's Method*

Newton's method is the basis of number of popular second order optimization algorithms. Newton's algorithm is iterative, where in each iteration, [2]

1. Calculate the Newton weight change vector **e**

2. Updates variable with this weight change vector, **e**

The weight change vector, **e** is calculated by solving the linear equations

$$\mathbf{H'e} = \mathbf{g'} \tag{2.33}$$

where, **H'** is the Hessian of the objective function calculated with respect to all the weights in the network and has elements defined as,

$$h'(i, j) = \frac{\partial^2 E}{\partial w(i) \partial w(j)} \tag{2.34}$$

**g'** is a vector of negative gradients and is defined as

$$\mathbf{g'} = vec(\mathbf{G}, \mathbf{G_{oh}}, \mathbf{G_{oi}}) \tag{2.35}$$

Once eq. (2.33) is solved for **e**, the weights are updated as

$$\mathbf{w} \leftarrow \mathbf{w} + \mathbf{e} \tag{2.36}$$

where, **w** is defined in eq. (2.28)



Even though Newton's method has fast convergence in terms of iterations, it may fail if its Hessian is singular [37]. In this case the Levenberg-Marquardt (LM) algorithm [35] is used.

*Levenberg-Marquardt Algorithm*

LM can be thought of as a combination of steepest descent and the Gauss-Newton method. When the current solution is far from the correct one, the algorithm behaves like a steepest descent method: slow, but guaranteed to converge. When the current solution is close to the correct solution, it becomes a Gauss-Newton method.

In LM the Gauss-Newton Hessian is modified as,

$$\mathbf{H'} \leftarrow \mathbf{H'} + \lambda \mathbf{I} \qquad (2.37)$$

where $\lambda$ is a constant and $\mathbf{I}$ is the Identity matrix of the same size as $\mathbf{H'}$. The weight change vector $\mathbf{e}$ is computed from eq. (2.38) using OLS [2] and the weights are updated as

$$\mathbf{e} = \left[\mathbf{H'} + \lambda \mathbf{I}\right]^{-1} \mathbf{g'} \qquad (2.38)$$

$$\mathbf{w} \leftarrow \mathbf{w} + \mathbf{e} \qquad (2.39)$$

By changing the value of $\lambda$ the algorithm can interpolate between first order and second order methods. If the error increases the value of $\lambda$ is increased and the algorithm mimics steepest descent and if the error decreases the value of $\lambda$ is decreased in order to operate it close to the Gauss-Newton method. LM is computationally very expensive, so it is used only for small networks [35].



Chapter 3

Practical Newton's Algorithms for the MLP

Newton's method uses first and second order derivatives and usually performs better than steepest descent in terms of convergence per iteration. The idea behind this method is that, given a starting point, a quadratic approximation to the objective function that matched the first and second derivative values at that point is constructed. Then the approximate quadratic error function to generate the current solution vector is minimized. The current solution is then the starting point in the next iteration. If the objective function is quadratic, then the approximation is exact, and the method yields the true minimum in one step. If, on the other hand, the objective function is not quadratic, then the approximation will provide only an estimate of the position of the true minimum [38]. Practical Newton's algorithms for the MLP are described in this section.

### 3.1 Output Weight Optimization

Output weight optimization (OWO) is a technique used to find output weights $\mathbf{W_o}$ [74], which include both output to hidden and bypass weights [$\mathbf{W_{oi}} : \mathbf{W_{oh}}$], with less computation than steepest descent. In this subsection we show that OWO is equivalent to applying Newton's method to output weights. The error function of (2.3) can be written as,

$$E = \frac{1}{N_v} \sum_{p=1}^{N_v} \sum_{i=1}^{M} \left( t_p(i) - \sum_{n=1}^{N+N_h+1} X_p(n) w_o(i,n) \right) \tag{3.1}$$

where, the basis vector $\mathbf{X_p}$ is the augmented input of size $N_u$ formed by combining actual input and hidden unit activations [$\mathbf{x_p}, \mathbf{O_p}$]$^{\mathrm{T}}$. In order to find the output weight change vector $\mathbf{d}$ calculation of first and second partial derivatives of the error function in eq. (3.1) with respect to the output weights is needed. The elements of the Hessian matrix $\mathbf{H_o}$ and the negative gradient vector $\mathbf{g_o}$ of output weights are computed as,



$$g_o(i,j) = \frac{-\partial E}{\partial w_o(i,j)} = \frac{2}{N_v} \sum_{p=1}^{N_v} \left( t_p(i) - \sum_{n=1}^{N+N_h+1} X_p(n) w_o(i,n) \right) X_p(j) \tag{3.2}$$

$$h_o(i,k,m,j) = \frac{\partial^2 E}{\partial w_o(i,k) \partial w_o(m,j)} = \frac{2}{N_v} \sum_{p=1}^{N_v} X_p(k) X_p(j)$$

$$\tag{3.3}$$

The elements of the 2 dimensional Hessian matrix $\mathbf{H_o}$, are found from $h_o(i,k,m,j)$ as

$$h_o\big((i-1)N_u + k, (m-1)N_u + j\big) = h_o(i,k,m,j) \tag{3.4}$$

where, $N_u = N_h + N + 1$. Elements of the negative gradient column vector $\mathbf{g_o}$ are calculated from $g_o(i,j)$ as,

$$g_o\big((i-1)N_u + j\big) = g_o(i,j) \tag{3.5}$$

The output weight matrix is converted into a vector $\mathbf{w_o} = \text{vec}\{\mathbf{W_o}\}$. The output weight change vector $\mathbf{d} = \mathbf{w_o}'^T - \mathbf{w_o}^T$, where $\mathbf{w_o}'$ is the new version of the output weight vector $\mathbf{w_o}$, is computed from Newton's method by solving the following equation.

$$\mathbf{H_o} \cdot \mathbf{d} = \mathbf{g_o} \tag{3.6}$$

Let us assume that there is only one output in the network. Now from eq. (2.23), (2.24), (3.2) and (3.3), we can see that

$$\mathbf{H_o} = 2 \cdot \mathbf{R} \ , \tag{3.7}$$

$$\mathbf{g_o} = 2 \cdot \big( \mathbf{C} - \mathbf{R w_o^T} \big) \tag{3.8}$$

By solving eq. (3.6), (3.7) and (3.8) we get

$$\mathbf{w'_o^T} - \mathbf{w_o^T} = \mathbf{R}^{-1} \big( \mathbf{C} - \mathbf{R w_o^T} \big) \tag{3.9}$$

$$\mathbf{w'_o^T} = \mathbf{R}^{-1} \mathbf{C} \tag{3.10}$$

This shows that OWO is nothing but Newton's method applied to output weights.



## 3.2 Newton's Method for Input Weights

Let the input weight matrix is converted into a vector, $\mathbf{w}=\text{vec}\{\mathbf{W}\}$. Let $\mathbf{e}=\mathbf{w'}-\mathbf{w}$ be the unknown input weight change vector, where $\mathbf{w'}$ is the new version of input weight vector $\mathbf{w}$ that we're trying to find. The multivariate Taylor's theorem says that E', which is E as a function of $\mathbf{e}$, can be approximated as [39],

$$E' \approx E - \mathbf{e}^T \cdot \mathbf{g} + \frac{1}{2}\mathbf{e}^T \cdot \mathbf{H} \cdot \mathbf{e} \tag{3.11}$$

where, the elements of the Gauss-Newton input weight Hessian matrix $\mathbf{H}$ can be computed as follows

$$h(k,n,j,m) = \frac{\partial^2 E}{\partial w(k,n)\partial w(j,m)} \tag{3.12}$$

$$= \frac{2}{N_v}\sum_{p=1}^{N_v}\sum_{i=1}^{M}\frac{\partial y_p(i)}{\partial w(k,n)}\frac{\partial y_p(i)}{\partial w(j,m)} \tag{3.13}$$

$$\frac{\partial y_p(i)}{\partial w(k,n)} = w_{oh}(i,k)f'\big(n_p(k)\big)x_p(n) \tag{3.14}$$

The elements of the 2 dimensional Hessian $\mathbf{H}$ in eq. (3.11) are found from h(k,n,j,m) as

$$h\big((k-1)N_h+n,(j-1)N_h+m\big) = h(k,n,j,m) \tag{3.15}$$

Elements of the negative gradient column vector $\mathbf{g}$ are calculated from elements of the negative gradient matrix $\mathbf{G}$ in eq. (2.10) as,

$$g\big((k-1)N_h+n\big) = g(k,n) \tag{3.16}$$

Setting $\partial E'/\partial \mathbf{e} = 0$, we get

$$\partial E'/\partial \mathbf{e} = -\mathbf{g} + \mathbf{H} \cdot \mathbf{e} = 0 \tag{3.17}$$

$$\mathbf{e} = \mathbf{H}^{-1} \cdot \mathbf{g} \tag{3.18}$$



The input weight vector is then updated as

$$\mathbf{w'} = \mathbf{w} + \mathbf{e} \tag{3.19}$$

### 3.3 OWO-Newton Algorithm

OWO-Newton is a two-step method in which we alternatively use Newton's algorithm to decrease E via OWO in section 3.1 and Newton's method for input weights in section 3.2 (2.22), (2.23) and (2.24). For the first iteration the input weights are randomly initialized. In the second step the input weights **W** are updated using Newton's algorithm [2]. This process involves computation of the input weight Hessian **H** and the negative input weight gradient matrix **G**, as shown in eq. (2.10), (3.12), (3.13) and (3.14). This information is used to compute input weight change vector **e** and in turn update the input weights as in eq. (3.18) and (3.19).

*OWO-Newton Algorithm*

Given the number of iterations, $N_{it}$

Initialize **W**

For $i_t$=1 to $N_{it}$

        Calculate **G**

        $\mathbf{g} \longleftarrow \text{vec}(\mathbf{G})$

        Calculate **H**

        Solve (3.3) for **e**

        $\mathbf{D} = \text{vec}^{-1}(\mathbf{e})$

        $\mathbf{W} \longleftarrow \mathbf{W} + \mathbf{D}$

        Perform OWO

end for

The OWO-Newton algorithm has the following problems.



1. The error is not guaranteed to always decrease, since E is not a quadratic function of **W**.

2. It is computationally very expensive since it involves computation and storage of Hessians and gradients; this makes it unfeasible for very large networks.

In order to reduce the computational cost, the OWO-MOLF algorithm for feedforward network was introduced [5].

## 3.4 OWO-MOLF Algorithm and its Analysis

In the OWO-BP algorithm the input weights are trained using the negative gradients and a learning factor z. The learning factor is normally assumed to be a small constant value; the learning factor z can also be found using a Taylor series as in eq. (2.16), (2.18), (2.19) and (2.20). The algorithm that uses the optimal learning factor performs extremely well when compared to arbitrary assumption. This shows that learning factor plays a very important role in the training process. In the OWO-MOLF algorithm the input weights of a network are trained using a negative gradient matrix **G** and a vector of learning factors **z** of size $N_h \times 1$. The derivation of multiple optimal learning factors algorithm is shown below [5] [75].

*Derivation of Multiple Optimal Learning Factors*

Consider an MLP with one hidden layer and $N_h$ hidden units. Let us also assume that this MLP is trained using OWO-BP. Consider a learning factor vector **z** with $N_h$ elements in it and all the input weights that are connected to a hidden unit are updated using the learning factor $z_k$ associated with that particular $k^{th}$ hidden unit. The error function to be minimized is given in eq. (2.3). The predicted output $y_p(i)$ is given by,

$$y_p(i) = \sum_{n=1}^{N+1} w_{oi}(i,n) x_p(n) + \sum_{k=1}^{N_h} w_{oh}(i,k) f\left(\sum_{n=1}^{N+1} (w(k,n) + z_k \ g(k,n)) x_p(n)\right) \quad (3.20)$$



where, $g(k,n)$ is an element of the negative Jacobian matrix **G**. The first partial derivative of $E$ with respect to $z_j$ is [5],

$$\frac{\partial E}{\partial z_j} = \frac{-2}{N_v} \sum_{p=1}^{N_v} \sum_{i=1}^{M} \left[ \bar{t}_p(i) - \sum_{k=1}^{N_h} w_{oh}(i,k) O_p(z_k) \right] w_{oh}(i,j) O'_p(j) \Delta n_p(j) \quad (3.21)$$

where,

$$\bar{t}_p(i) = t_p(i) - \sum_{n=1}^{N+1} w_{oi}(i,n) x_p(n), \qquad \Delta n_p(j) = \sum_{n=1}^{N+1} x_p(n) g(j,n), \quad (3.22)$$

$$O_p(z_k) = f\left( \sum_{n=1}^{N+1} \left( w(k,n) + z_k g(k,n) \right) x_p(n) \right) \quad (3.23)$$

Using Gauss-Newton updates, the second partial derivative elements of the Hessian **H$_{molf}$** are

$$h_{molf}(k,j) \approx \frac{\partial^2 E}{\partial z_k \partial z_j} = \frac{2}{N_v} \sum_{p=1}^{N_v} \sum_{i=1}^{M} \frac{\partial y_p(i)}{\partial z_k} \frac{\partial y_p(i)}{\partial z_j} \quad (3.24)$$

$$\frac{\partial y_p(i)}{\partial z_k} = w_{oh}(i,k) f'\left(n_p(k)\right) \sum_{n=1}^{N+1} x_p(n) g(k,n) \quad (3.25)$$

$$h_{molf}(k,j) = \sum_{m=1}^{N+1} \sum_{n=1}^{N+1} \left[ \frac{2}{N_v} \sum_{i=1}^{M} w_{oh}(i,k) w_{oh}(i,j) \sum_{p=1}^{N_v} x_p(m) x_p(n) O'_p(k) O'_p(j) \right] g(k,m) g(j,n)$$

$$(3.26)$$

The Gauss-Newton update guarantees that **H$_{molf}$** is non-negative definite. Given the negative gradient vector, $\mathbf{g_{molf}} = \left[ -\partial E / \partial z_1, -\partial E / \partial z_2 \ldots, -\partial E / \partial z_{N_h} \right]^T$ and the Hessian **H$_{molf}$**, the next step is to minimize E with respect to the vector **z** using Newton's method. Note that -g$_{molf}$(j) is given in (3.21), (3.22) and (3.23). In each iteration of the OWO-MOLF algorithm, the steps are as follows [5]:

1. Calculate the negative input weight Jacobian **G** using BP.



2. Calculate **z** using Newton's method and update the input weights as

$$w(k, n) \leftarrow w(k, n) + z_k g(k, n) \tag{3.27}$$

3. Solve linear equations for all output weights.

Here, the MOLF procedure has been inserted into the OWO-BP algorithm, resulting in an algorithm we denote as OWO-MOLF. The MOLF procedure can be inserted into other algorithms as well.

*MOLF Hessian from Gauss-Newton's Hessian*

If $\mathbf{H_{molf}}$ and $\mathbf{g_{molf}}$ are the Hessian and negative gradients, respectively of the error with respect to **z**, then the multiple optimal learning factors are found by solving

$$\mathbf{H_{molf} z = g_{molf}} \tag{3.28}$$

The term within the square bracket in (3.26) is the elements of the Hessian **H** from Gauss-Newton method for updating input weights (as in (3.17), (3.18) and (3.19)). Hence,

$$\frac{\partial^2 E}{\partial z_k \partial z_j} = \sum_{m=1}^{N+1} \sum_{n=1}^{N+1} \left[ \frac{\partial^2 E}{\partial w(k, m) \partial w(j, n)} \right] g(k, m) g(j, n) \tag{3.29}$$

*Lemma 4.* The $(k,j)^{\text{th}}$ elements of $\mathbf{h_{molf}}$ can be expressed in vector notation as,

$$\frac{\partial^2 E}{\partial z_k \partial z_j} = \sum_{m=1}^{N+1} g(k, m) \sum_{n=1}^{N+1} h^{k,j}(m, n) g(j, n) \tag{3.30}$$

In matrix notation,

$$h_{molf}(k, j) = \mathbf{g_k^T H^{k,j} g_j} \tag{3.31}$$

where, the column vector $\mathbf{g_k}$ contains **G** elements $g(k,n)$ for all values of n, where the (N+1) by (N+1) matrix $\mathbf{H^{k,j}}$ contains elements of **H** for weights connected to $k^{\text{th}}$ and $j^{\text{th}}$ hidden units.



*Analysis of OWO-MOLF*

OWO-MOLF algorithm is very attractive for various reasons. First, it is not computationally expensive due to the reduced size of the Hessian. Second, linear dependencies in input and hidden unit activations do not lead to the singularity of the MOLF Hessian. This method performs as well as or better than the Levenberg-Marquardt, with several orders of magnitude fewer multiplies. The MOLF hessian and gradients are weighted sums of the total network Hessian and gradients; this interesting property makes it very flexible to switch from the OWO-Newton algorithm to the OWO-MOLF algorithm.

In MOLF the number of learning factors that are computed is always equal to the number of hidden units $N_h$ present in the network. The adaptive MOLF algorithm introduced in chapter 5, adapts the dimension of the learning factor vector **z**.

### 3.5 Computational Analyses of Two-Step Algorithms

As discussed earlier the OWO technique involves computation of the autocorrelation of the augmented input vector **$X_p$** and the cross correlation matrix of the augmented input vector with the desired output vector **$t_p$**. OLS is used to solve the set of linear equations in eq. (2.22) in order to find the output weights **$W_o$**. The number of multiplies required to solve for the output weights using OLS in one iteration is given by,

$$M_{ols} = N_u \left( N_u + 1 \right) \left[ M + \frac{1}{6} \left( 2N_u + 1 \right) + \frac{3}{2} \right] \tag{3.32}$$

The total number of multiplies required to compute the output weights and negative gradients in one iteration is given by [5],

$$M_{OWO\text{-}BP} = N_u \left( N_u + 1 \right) \left[ M + \frac{1}{6} \left( 2N_u + 1 \right) + \frac{3}{2} + \frac{N_v}{2} \right] + N_v \left[ N_h \left( M + 2N + 3 \right) + M \left( 2N_u + 1 \right) \right] \tag{3.33}$$



The number of multiplies per iteration of LM is given below [5],

$$M_{lm} = N_v[MN_u + 2N_h(N+1) + M(N + 6N_h + 4) + MN_u(N_u + 3N_h(N+1)) +$$

$$4N_h^2(N+1)^2] + N_w^3 + N_w^2 \qquad (3.34)$$

where, $N_w$ is the total number of weights in the network and is equal to $MN_u+(N+1)N_h$ and $N_u=N+N_h+1$

Newton's method for input weights involves computation of the input weight Hessian matrix of size $N_{iw} \times N_{iw}$, where $N_{iw}=N_h(N+1)$. The number of multiplies required for one iteration is given as

$$M_{Newton} = N_v\left[N_{iw}(2M+1) + N_{iw}(N_{iw}+1)\left[\frac{N_v M}{2} + \frac{1}{6}(2N_{iw}+1) + \frac{5}{2}\right]\right] \quad (3.35)$$

In the OWO-Newton algorithm, the total number of multiplies required per iteration is,

$$M_{OWO-Newton} = M_{OWO} + M_{Newton} \qquad (3.36)$$

where,

$$M_{OWO} = N_v\left[N_h(N+1) + M(2N_u+1)\right] + N_u(N_u+1)\left[M + \frac{N_v}{2} + \frac{1}{6}(2N_u+1) + \frac{3}{2}\right](3.37)$$

The OWO-MOLF algorithm also involves computation of a Hessian, however, compared to Newton's method for input weights or LM, the size of the Hessian is much smaller. The Hessian used in OWO-MOLF has only $N_h$ rows and columns. The number of multiplies required for OWO-MOLF in one iteration is given as,

$$M_{OWO-MOLF} = N_h(N_h+1)\left[\frac{1}{6}(2N_h+1) + \frac{5}{2}\right] + N_v N_h\left[2M + N + 2 + \frac{M(N_h+1)}{2}\right]$$

$$(3.38)$$



Chapter 4

Motivation for Additional Work

4.1 Problems with existing training algorithms

Attempts to develop second order training algorithms have fallen short because of the following problems

1. We have no theoretical justification for introducing small second order modules into first order algorithms as is done in OWO-MOLF.

2. Second order training is computationally very expensive when there are many unknowns as in LM and OWO-Newton.

3. Algorithms such as OWO-MOLF may obtain more error decrease per multiply than LM and OWO-Newton. The latter two algorithms provide more error decrease per iteration. We lack a method for interpolating between OWO-MOLF and OWO-Newton

4.2 Proposed Work

We propose to solve the listed problems with the following tasks:

1. Show that in theory, incrementally increasing the size of the unknown vector in second order modules helps.

2. Develop a method for changing the size of second order training modules for optimizing error decrease per multiply

3. Develop a method that interpolates between OWO-MOLF and OWO-Newton.



## 4.3 Relevent Lemmas

Here I present Lemma 1 in order to perform task 1

*Lemma 1*:

Assume E($\mathbf{w}$) is a quadratic function of the input weight vector $\mathbf{w}$ which is divided into k partitions $\mathbf{w_k}$ such that $\mathbf{w} = \left[\mathbf{w_1^T}; \mathbf{w_2^T}; \cdots \mathbf{w_k^T}\right]^T$ and $\mathbf{g_k} = \dfrac{\partial E}{\partial \mathbf{w_k}}$. If a training algorithm minimizes E with respect to the k dimensional vector $\mathbf{z}$ producing an error $E_k = E\left(\mathbf{w_1} + z_1\mathbf{g_1}, \mathbf{w_2} + z_2\mathbf{g_2}, \dots \mathbf{w_k} + z_k\mathbf{g_k}\right)$ and k can only increase by splitting one of the existing partitions, then $E_{k+1} \le E_k$

*Proof*:

The error E($\mathbf{w}$) after updating the input weights can be modeled as,

$$E(\mathbf{w} + \mathbf{e}) = E_o - \mathbf{e^T g} + \frac{1}{2}\mathbf{e^T H e} \tag{4.1}$$

where $E_o$ is the error before updating the input weights, $\mathbf{g}$ is $\mathbf{g_k}$ for k=1, $\mathbf{H}$ is the Hessian, and $\mathbf{e}$ is the input weight change vector. If $\mathbf{e}$ is found optimally using Newton's method, then

$$\mathbf{e} = \mathbf{H^{-1}} \cdot \mathbf{g} \tag{4.2}$$

The input weight change vector for k groups is

$$\mathbf{e_k} = \left[z_1\mathbf{g_1^T}, z_2\mathbf{g_2^T} \dots z_k\mathbf{g_k^T}\right]^T \tag{4.3}$$

Given, $\mathbf{z} = \text{argmin}_z\left(E(\mathbf{w} + \mathbf{e_k})\right)$, increase k by one so that

$$\mathbf{e_{k+1}} = \left[z_1\mathbf{g_1^T}, z_2\mathbf{g_2^T} \dots z_{ka}\mathbf{g_{ka}^T}, z_{kb}\mathbf{g_{kb}^T}\right]^T \tag{4.4}$$

If $z_{ka} = z_{kb} = z_k$, then $\mathbf{e_k} = \mathbf{e_{k+1}}$ and $E_{k+1} = E_k$. However since the k+1 elements in $\mathbf{z}$ can all change, we get $E_{k+1} \le E_k$



Lemma 1 presents a clear justification for increasing the number of second order groups or unknowns

Task 2 is performed by developing new algorithm of adaptive multiple optimal learning factors which is explained in detail in chapter 5

In OWO-MOLF all the input weights connected to a hidden unit forms a group. $N_h$ learning factors are computed one for each group. In adaptive MOLF k groups of input weights are formed by splitting the $N_h$ groups of OWO-MOLF. Here I present lemma 2 and 3 in order to show that the proposed algorithm of adaptive multiple optimal learning factors does perform task 3.

Let adaptive MOLF denote any algorithm in which the weight groups of OWO-MOLF have been split, generating new groups and therefore increasing k in lemma 1.

*Lemma 2*:

If E($\mathbf{w}$) is quadratic in a given iteration, and if E-E$_{MOLF}$ and E-E$_{aMOLF}$ denote the error decrease due to the Newton steps of OWO-MOLF and adaptive MOLF respectively, then E-E$_{MOLF}$ ≤ E-E$_{aMOLF}$.

*Proof*:

The k groups of unknowns for adaptive MOLF can be formed by splitting the $N_h$ groups of OWO-MOLF. The lemma follows from Lemma1.

*Lemma 3*:

OWO-Newton is a limiting case of the adaptive MOLF algorithm as the k groups of adaptive MOLF are split until k=$N_h$·(N+1)

*Proof*:



We have
$$\mathbf{e_{Newton}} = \begin{bmatrix} z_1 \cdot g_1 \\ z_2 \cdot g_2 \\ z_3 \cdot g_3 \\ \bullet \\ \bullet \\ z_{Nh(N+1)} \cdot g_{Nh(N+1)} \end{bmatrix}$$

(4.5)

In a given iteration's Newton steps, the resulting errors are related as

$E_{Newton} \leq E_{aMOLF}$. Equality occurs if $g_k \neq 0$ for every non zero element of $e_{Newton}$.

Lemmas 2 and 3 show that the adaptive MOLF algorithm interpolates between OWO-MOLF and OWO-Newton.



Chapter 5

Adaptive Multiple Optimal Learning Factors Algorithm

5.1 Core Idea of Adaptive MOLF Algorithm

The OWO-Newton algorithm often has excellent error convergence characteristics, but its performance decreases when there are linear dependencies in the input signal or hidden unit activations, and when the error function is not well approximated by a quadratic function. Also OWO-Newton is computationally more expensive than first order algorithms. Although the OWO-MOLF algorithm is not as good as OWO-Newton in convergence per iteration, it is much more resistant to linearly dependent input signals and it is computationally less expensive. The main idea of the Adaptive MOLF algorithm is to preserve the advantages of both OWO-MOLF and OWO-Newton and to get rid of their drawbacks.

This algorithm can adapt itself between OWO-MOLF and OWO-Newton algorithms. In OWO-MOLF, $N_h$ learning factors are computed. In the adaptive MOLF algorithm the number of learning factors that are computed can be anywhere between $N_h$ and $N_h \times (N+1)$ in each iteration.

5.2 Grouping of the Input Weights

In the algorithm being described, a group is a set of input weights that are updated using the same learning factor. In the steepest descent method for example, there is one learning factor, z, and the group contains all network weights.

In the OWO-MOLF algorithm all the input weights that are connected to a hidden unit are updated using a learning factor associated with that particular hidden unit. In this case the group size is N+1 and the total number of groups per hidden unit, $N_g$ is equal to one. In the adaptive MOLF algorithm the group size changes between N+1 and 1 and the number of groups per hidden unit varies between 1 and N+1.



In this algorithm the input weights **W** are grouped based on the curvature $\mathbf{H_w}$ of the error function calculated with respect to input weights (the second derivative of the error function in eq. (2.3) with respect to the input weights). The elements of the matrix $\mathbf{H_w}$ can be calculated as,

$$h_w(k,n) = \frac{\partial^2 E}{\partial w(k,n)^2} = \frac{2}{N_v} \sum_{i=1}^{M} w_{oh}(i,k)^2 \sum_{p=1}^{N_v} f'(n_p(k))^2 x_p(n)^2$$

(5.1)

Weights that are connected to different hidden units are never grouped since that is observed to damage performance.

### 5.3 Adapting the number of groups

Our goal is to vary the number of groups per hidden unit so that the error change per multiply is maximized. In each iteration, error change is computed by taking the difference in the error from present and previous iterations. This error change is divided by the number of multiples required for the current iteration as shown in eq. (5.2) in order to obtain the error change per multiply (EPM) attained in the present iteration.

$$EPM(i_t) = \frac{E(i_t - 1) - E(i_t)}{M(i_t)}$$

(5.2)

where, $M(i_t)$ stands for number of multiplies in iteration $i_t$

$EPM(i_t)$ is the error per multiply in iteration $i_t$

$E(i_t)$ it the error computed in iteration $i_t$

$i_t$ is the current iteration number.

The number of groups per hidden unit $N_g$ increase as the error change per multiply increases and vice versa. This algorithm creates a large error decrease in the initial iterations by operating similar to the OWO-Newton algorithm and as the error starts converging to the local minima the algorithm adapts itself to operate close to the OWO-MOLF algorithm in order to reduce the computational burden.



## 5.4 Derivation of the Adaptive MOLF algorithm

Assume an MLP whose input weights are trained using negative gradients as shown in section 2.2.1. But instead of a single learning factor assume a vector of learning factors $\mathbf{z}$, with elements $z_{k,C}$ used to update all the weights $w(k,n)$ that belong to group C of hidden unit k. The error function to be minimized is given in eq. (2.3). The predicted output $y_p(i)$ is given by

$$y_p(i) = \sum_{n=1}^{N+1} x_p(n) w_{oi}(k,n) + \sum_{k=1}^{N_h} w_{oh}(i,k) f\left( \sum_{C=1}^{N_g} \sum_{a=R(C-1)+1}^{R(C)} x_p(i_k(a)) \left[ w(k,i_k(a)) + z_{k,C} g(k,i_k(a)) \right] \right)$$

(5.3)

where,

$$R(C) = C \cdot Gs(C)$$

(5.4)

$$Gs(0) = 0$$

(5.5)

where,   $N_g$ is the number of groups per hidden unit

C is the group index

$\mathbf{Gs}$ is the array of $N_g \times 1$ elements, these elements contains the sizes of each group, i.e. number of inputs associated to each group.

$\mathbf{I_k}$ is the vector of input indices ordered in such a way that the index n of input to which weight $w(k,n)$ with higher curvatures are connected will come first.

$\mathbf{I_k} = [n_1, n_2, n_3, \ldots, n_{N+1}]$ where $n_1, n_2, n_3, \ldots, n_{N+1}$ are input indices such that

$h_w(k,n_1) \geq h_w(k,n_2) \geq h_w(k,n_3) \ldots \ldots \geq h_w(k,n_{N+1})$

$z_{k,C}$ is the learning factor used to update all the input weights that belong to group C of hidden unit k.

R is the function that maps between the group indices and the indices of vector $\mathbf{I}$



The total number of learning factors that are going to be computed is $L = N_h \cdot N_g$. $L$ varies from $N_h$ to $N_h \cdot (N+1)$ as the number of groups per hidden unit $N_g$, changes from 1 to $N+1$. The first negative partial of E with respect to $z_{k,C}$ is,

$$g_{Amolf}(k,C) = \frac{\partial E}{\partial z_{k,C}} = \frac{2}{N_v} \sum_{p=1}^{N_v} \sum_{i=1}^{M} (t_p(i) - y_p(i)) \frac{\partial y_p(i)}{\partial z_{k,C}}$$

(5.6)

where,

$$\frac{\partial y_p(i)}{\partial z_{k,C}} = w_{oh}(i,k) f'(n_p(k)) \Delta n_p(k,C)$$

(5.7)

$$n_p(k) = \sum_{C=1}^{NClust} \sum_{a=R(C-1)+1}^{R(C)} x_p(i_k(a)) [w(k,i_k(a)) + z_{k,C} g(k,i_k(a))]$$

(5.8)

$$\Delta n_p(k,C) = \sum_{a=R(C-1)+1}^{R(C)} x_p(i_k(a)) g(k,i_k(a))$$

(5.9)

Using Gauss-Newton updates, the elements of the 4-D Hessian $\mathbf{H_{Amolf}}$ are computed as,

$$h_{Amolf}(k,C_1,j,C_2) = \frac{2}{N_v} \sum_{i=1}^{M} w_{oh}(i,k) w_{oh}(i,j) \sum_{p=1}^{N_v} f'(n_p(k)) f'(n_p(j)) \Delta n_p(k,C_1) \Delta n_p(j,C_2)$$

(5.10)

$$= \sum_{a=R(C_1-1)+1}^{R(C_1)} \sum_{b=R(C_2-1)+1}^{R(C_2)} \left[ \frac{2}{N_v} \sum_{i=1}^{M} w_{oh}(i,k) w_{oh}(i,j) \sum_{p=1}^{N_v} x_p(i_k(a)) x_p(i_j(b)) f'(n_p(k)) f'(n_p(j)) \right]$$
$$g(k,i_k(a)) g(j,i_j(b))$$

(5.11)

The elements of the 2 dimensional Hessian $\mathbf{H_{Amolf}}$ are found from $h_{Amolf}(k,C_1,j,C_2)$ as,

$$h_{Amolf}((k-1)N_g + C_1, (j-1)N_g + C_2) = h_{Amolf}(k,C_1,j,C_2)$$

(5.12)

Elements of negative gradient column vector $\mathbf{g_{Amolf}}$ are calculated from $g_{Amolf}(k,C)$ as,

$$g_{Amolf}((k-1)N_g + C) = g_{Amolf}(k,C)$$

(5.13)



The Gauss-Newton update guarantees that $\mathbf{H_{Amolf}}$ is non-negative definite. Given the negative gradient vector, $\mathbf{g_{Amolf}} = \left[ \dfrac{-\partial E}{\partial z_{1,C_1}}, \dfrac{-\partial E}{\partial z_{1,C2}} \ldots\ldots, \dfrac{-\partial E}{\partial z_{N_h,C_{N_g}}} \right]$ and the Hessian $\mathbf{H_{Amolf}}$, we minimize E with respect to vector $\mathbf{z}$ using Newton's method. The learning factors $\mathbf{z}$ can be computed as,

$$\mathbf{z} = \mathbf{H_{Amolf}^{-1}} \cdot \mathbf{g_{Amolf}} \tag{5.14}$$

The input weights $\mathbf{W}$ are updated as follows,

$$w\big(k, i_k(a)\big) = w\big(k, i_k(a)\big) + z_{k,C} g\big(k, i_k(a)\big) \tag{5.15}$$

where, $[(C-1)\cdot Gs(C-1)] + 1 \leq a \leq C \cdot Gs(C)$

*Initial point of the algorithm*

In sections 5.1, 5.2, 5.3 and 5.4 we have seen that the algorithm adapts itself between OWO-MOLF and OWO-Newton algorithm as the number of groups per hidden unit changes from N+1 to 1. Now in this section we discuss the starting point of the algorithm i.e., the number of groups per hidden unit that are chosen in first iteration.

This process is done by experimenting on all possibilities of the number of groups per hidden unit from 1 to N+1. The one that gives the lowest error is chosen to be the starting point. This technique improves the performance of the algorithm drastically compared to random initialization of the number of groups in the first iteration. This technique can be used at regular intervals in training say once in every 50 iterations to improve the performance.

Testing the error for all the possible numbers of groups is computationally expensive. This problem is solved by taking advantage of the concept that the Hessian and negative gradients of the Adaptive MOLF algorithm can be interpolated from Gauss-Newton Hessian and negative gradients of input weights.



*Adaptive MOLF Hessian and Gradients*

From eq. (5.6), (5.11), (3.11) and (2.10) it can be seen that the Hessian and negative gradients of Adaptive MOLF algorithm can be interpolated from Gauss-Newton input weight Hessian and negative gradients $\mathbf{H}$ and $\mathbf{g}$ respectively as,

$$g_{Amolf}(k, C) = \sum_{a=R(C-1)+1}^{R(C)} g(k, i_k(a))^2$$

(5.16)

$$h_{Amolf}(k, C_1, j, C_2) = \sum_{a=R(C_1-1)+1}^{R(C_1)} \sum_{b=R(C_2-1)+1}^{R(C_2)} h(k, i_k(a), j, i_j(b)) g(k, i_k(a)) g(j, i_j(b))$$

(5.17)

From eq. (5.14) and (5.15) we can interpolate the Hessian and negative gradients of an adaptive MOLF for any number of groups from Newton's Hessian and negative gradients of input weights. This helps to avoid recalculating Hessian and negative gradients for all the possible number of groups in determining initial point of the algorithm.

### 5.5 Effects of linearly dependent input signals

Let us assume an input signal $x_p(N+2)$ that is linearly dependent on other inputs as,

$$x_p(N+2) = \sum_{n=1}^{N+1} b(n) x_p(n)$$

(5.18)

During input weight adaptation the expression for negative gradient with respect to the learning factor associated to the group $C_1$ containing $x_p(N+2)$ is given as,

$$\frac{\partial E}{\partial z_{k,c_1}} = \frac{2}{N_v} \sum_{p=1}^{N_v} \sum_{i=1}^{M} (t_p(i) - y_p(i)) w_{oh}(i, k) f'(n_p(k)) [\Delta n_p(k, C_1) + x_p(N+2) g(k, N+2)]$$

(5.19)



$$= \frac{2}{N_v} \sum_{p=1}^{N_v} \sum_{i=1}^{M} \left(t_p(i) - y_p(i)\right) w_{oh}(i,k) f'(n_p(k)) \left[\Delta n_p(k,C_1) + \left[\sum_{n=1}^{N+1} x_p(n) b(n)\right] g(k,N+2)\right]$$

(5.20)

The elements of the Hessian computed with respect to learning factors associated with groups $C_1$ and $C_2$ which contain $x_p(N+2)$ is given as,

$$\overline{h_{Amolf}}(k,C_1,j,C_2) = \frac{\partial^2 E}{\partial z_{k,c_1} \partial z_{j,c_2}} = \frac{2}{N_v} \sum_{i=1}^{M} w_{oh}(i,k) w_{oh}(i,j) \sum_{p=1}^{N_v} f'(n_p(k)) f'(n_p(j)) [\Delta n_p(k,C_1) \Delta n_p(j,C_2)$$

$$+ g(k,N+2) \Delta n_p(j,N+2) \sum_{n=1}^{N+1} b(n) x_p(n) + g(j,N+2) \Delta n_p(k,N+2) \sum_{s=1}^{N+1} b(s) x_p(s)$$

$$+ g(k,N+2) g(j,N+2) \sum_{s=1}^{N+1} \sum_{n=1}^{N+1} b(n) x_p(n) b(s) x_p(s)]$$

(5.21)

$$\overline{h_{Amolf}}(k,C_1,j,C_2) = h_{Amolf}(k,C_1,j,C_2) + \frac{2}{N_v} \sum_{i=1}^{M} w_{oh}(i,k) w_{oh}(i,j) \sum_{p=1}^{N_v} x_p(N+2) f'(n_p(k)) f'(n_p(j))$$

$$\left[g(k,N+2) \Delta n_p(j,N+2) + g(j,N+2) \Delta n_p(k,N+2) + g(k,N+2) g(j,N+2) x_p(N+2)\right]$$

(5.22)

Comparing eq. (5.6) and (5.20), (5.10) and (5.22) it is clear that the cross terms shown in the square brackets in eq. (5.20) and (5.22) avoids the elements of the Hessian and gradient in Adaptive MOLF algorithm from being linearly dependent on others. But as the number of groups increases these cross terms disappear making few rows of $\overline{\mathbf{H_{Amolf}}}$ dependent on others, therefore making $\overline{\mathbf{H_{Amolf}}}$ singular. This can be prevented by just avoiding the number of groups per hidden unit $N_g$ to reach very close to N+1.



## 5.6 Computational Cost

The number of multiples required for an iteration of an adaptive MOLF algorithm is

$$M_{adaptive\_MOLF} = N_L\left(N_L + 1\right)\left[\frac{2N_L + 1}{6} + \frac{5}{2} + \frac{MN_v}{2}\right] + N_h\left(N + 1\right) + N_h N_g M\left(N_v + 2\right) + N_L N_v M$$

(5.23)

where, $N_L = N_g \cdot N_h$. The computational cost of adaptive MOLF algorithm also varies between computational cost of OWO-MOLF and OWO-Newton.



Chapter 6

Simulations

In this chapter the peformances of adaptive MOLF, OWO-MOLF, Scaled Conjugate Gradient and Levnberg Marquardt algorithms are compared. All the simulations shown in this chapter are run in Matlab 2013.

6.1 Experimental Plots

The performance of all the above mentioned algorithms is measured and compared both in terms of the number of iterations and the number of multiplies. In the LM and SCG algorithms all the weights are varied in every iteration. In the OWO-MOLF and adaptive MOLF algorithms the input weights are updated first and subsequently linear equations for the output weights are solved.

The data sets used for the simulations are listed below in Table 5.1. A detailed description of the different datasets is specified in Appendix A.

Table 6.1 Data Set Description

| Data Set Name | No. of Inputs | No. of Outputs | No. of Patterns |
|---------------|---------------|----------------|-----------------|
| Twod.tra | 8 | 7 | 1768 |
| Single2.tra | 16 | 3 | 10000 |
| Oh7.tra | 20 | 3 | 15000 |
| Concrete Data Set | 8 | 1 | 730 |
| Matrix Inversion | 4 | 4 | 2000 |

The above data sets are normalized to zero mean before training. The number of hidden units to be used in the MLP is determined by network pruning using the method of [75]. By this process the complexity of each of the data sets is analyzed and an appropriate number of hidden units is found. Training is done on the entire data set 10 times with 10 different initial networks. The average Mean Squared Error (MSE) from this 10-fold training is shown in the plots below.



The average training error and the number of multiplies is calculated for every iteration in a particular dataset using the different training algorithms. These results are then plotted to provide a graphical representation of the efficiency and quality of the different training algorithms. These plots for different datasets are shown below.

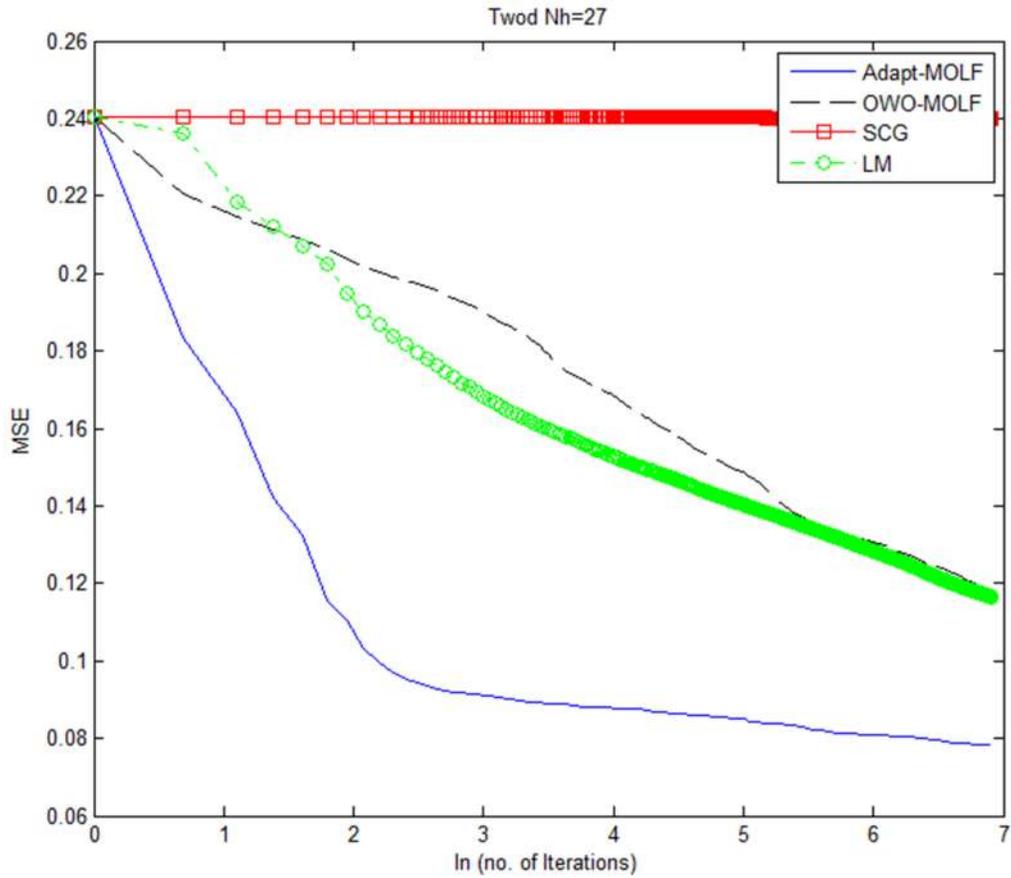

Figure 6.1 Twod.tra data set: Average error vs. number of iterations



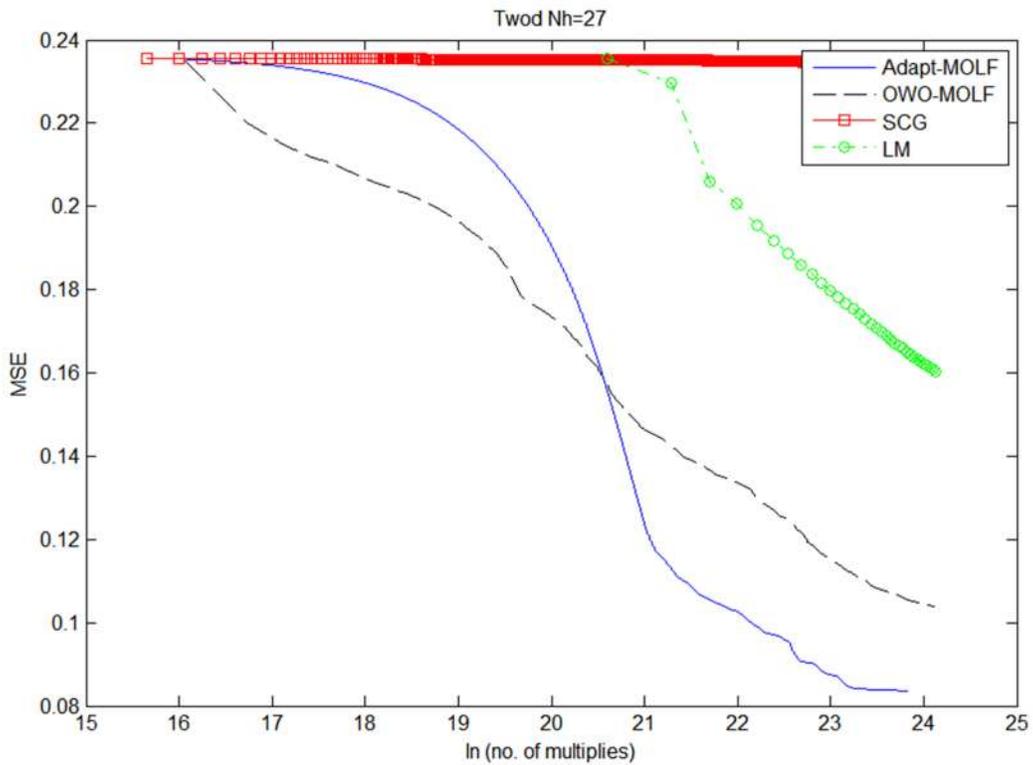

Figure 6.2 Twod.tra data set: average error vs. number of multiplies

For the Twod.tra data file [76], the MLP is trained with 27 hidden units. In Figure 6.1, the average mean square error (MSE) from 10-fold training is plotted versus the number of iterations for each algorithm (shown on a $\log_e$ scale). In Figure 6.2, the average training MSE from 10-fold training is plotted versus the required number of multiplies (shown on a $\log_e$ scale). From these plots it is clear that adaptive MOLF is performing far better than the other three algorithms both in terms of number of iterations and number of multiplies.



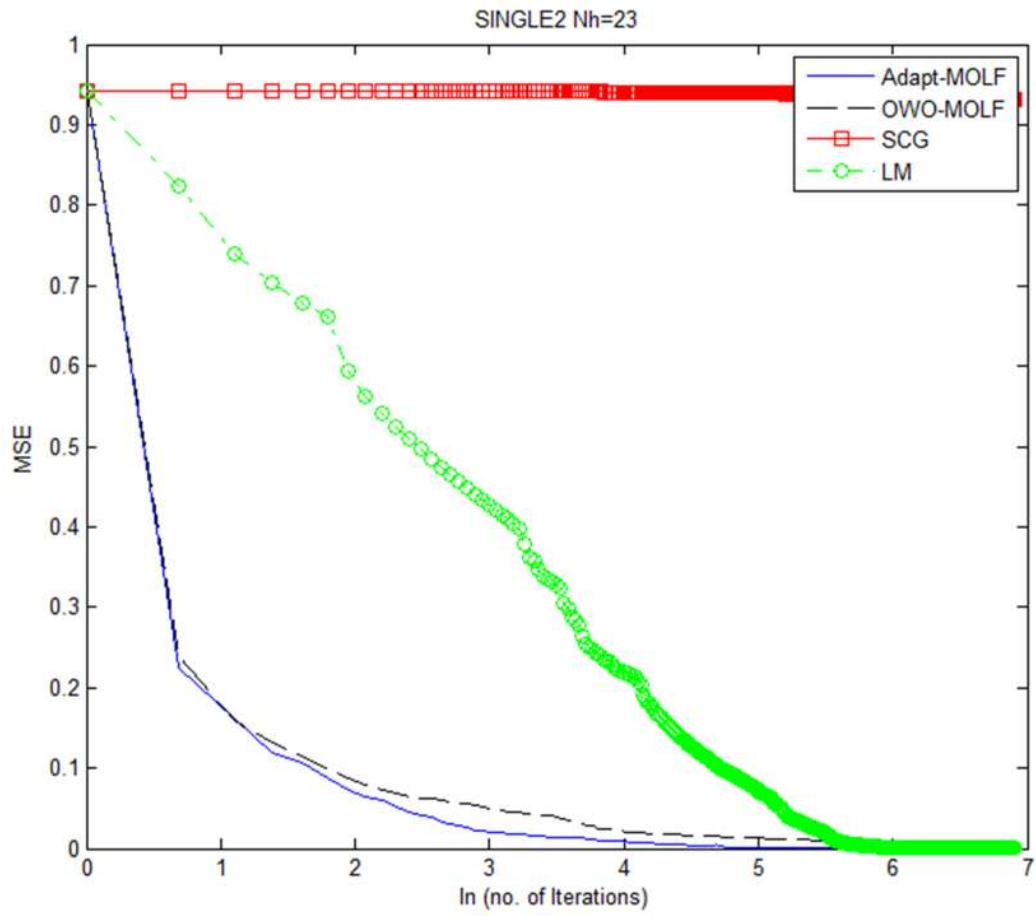

Figure 6.3 Single2.tra data set: average error vs. number of iterations



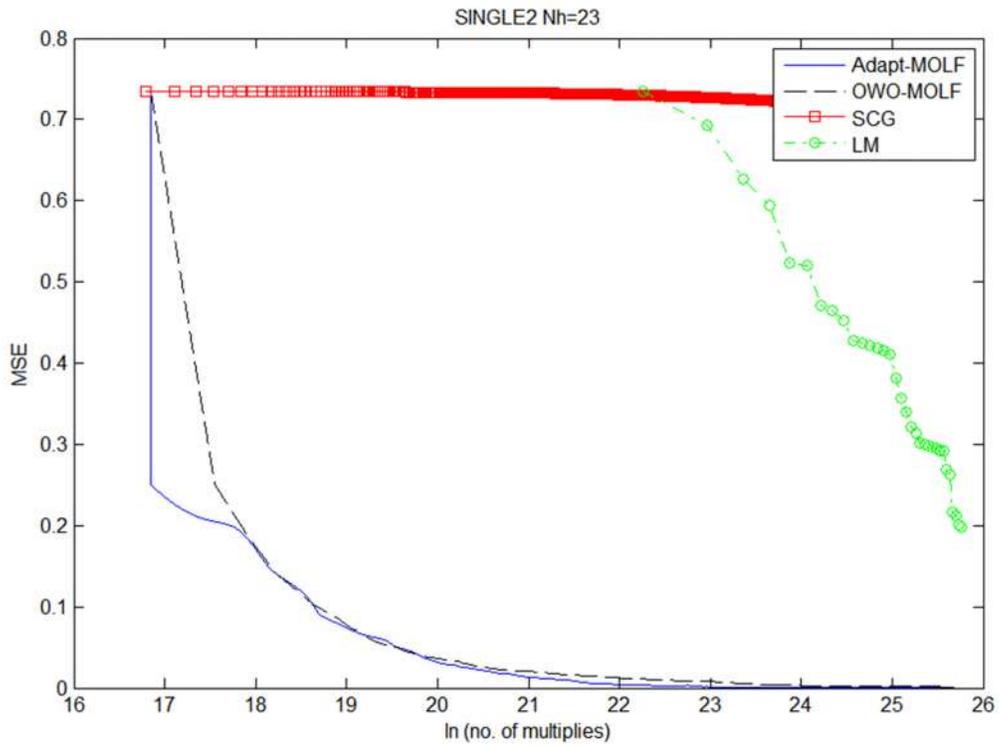

Figure 6.4 Single2.tra data set: average error vs. number of multiplies

For the Single2.tra data file [76], the MLP is trained with 23 hidden units. In Figure 6.3, the average mean square error (MSE) from 10-fold training is plotted versus the number of iterations for each algorithm (shown on a $\log_e$ scale). In Figure 6.4, the average training MSE from 10-fold training is plotted versus the required number of multiplies (shown on a $\log_e$ scale). For this dataset the performance of adaptive MOLF is very close to that of OWO-MOLF. This shows that the number of groups per hidden unit $N_g$ is 1 in most of the iterations.



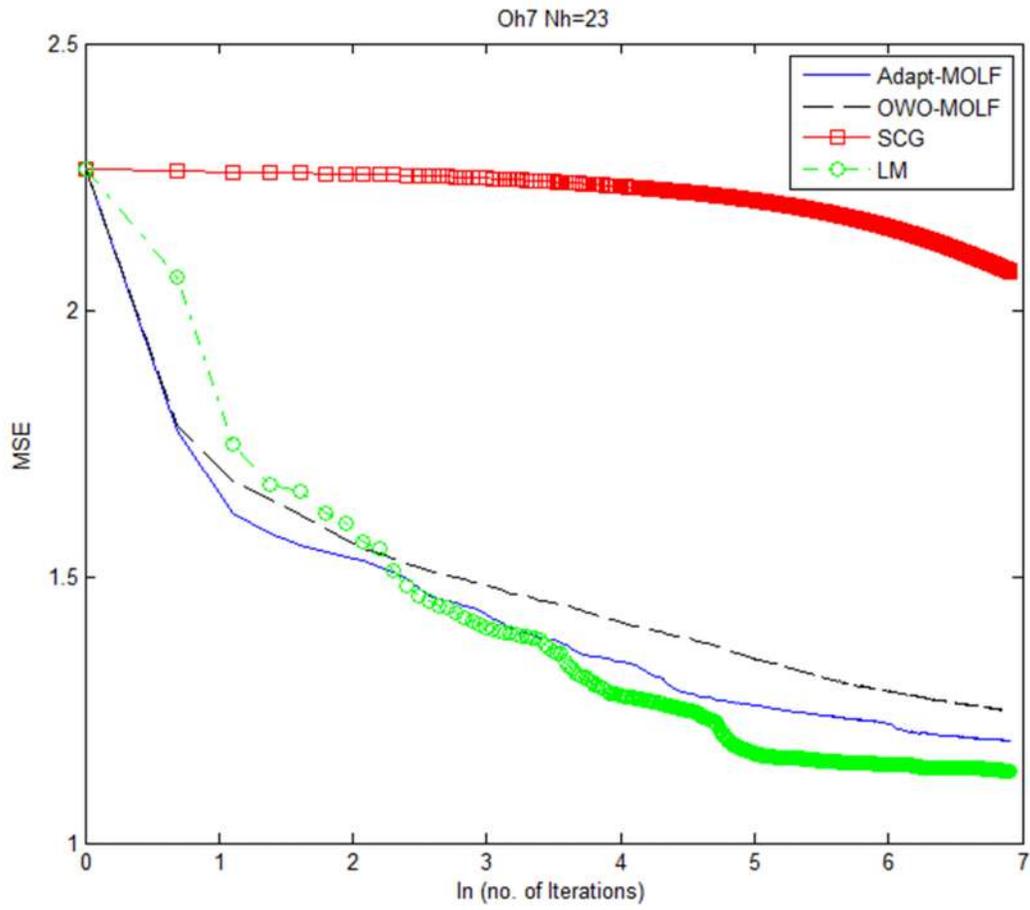

Figure 6.5 Oh7.tra data set: average error vs. number of iterations



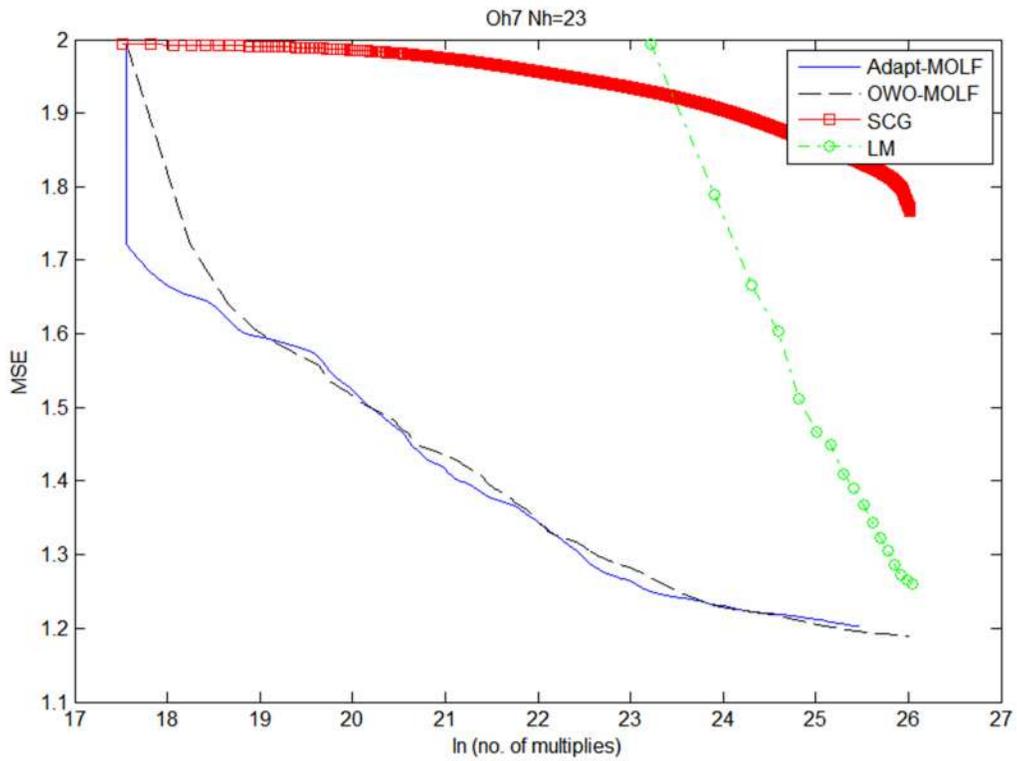

Figure 6.6 Oh7.tra data set: average error vs. number of multiplies

For the Oh7.tra data file [76], the MLP is trained with 23 hidden units. In Figure 6.5, the

average mean square error (MSE) for training from 10-fold training is plotted versus the

number of iterations for each algorithm (shown on a $\log_e$ scale). In Figure 6.6, the

average training MSE from 10-fold training is plotted versus the required number of

multiplies (shown on a $\log_e$ scale). In this data set the proposed algorithm is performing

better than OWO-MOLF in terms of iterations. In terms of multiplies the performance of

adaptive MOLF is very close to that of OWO-MOLF.



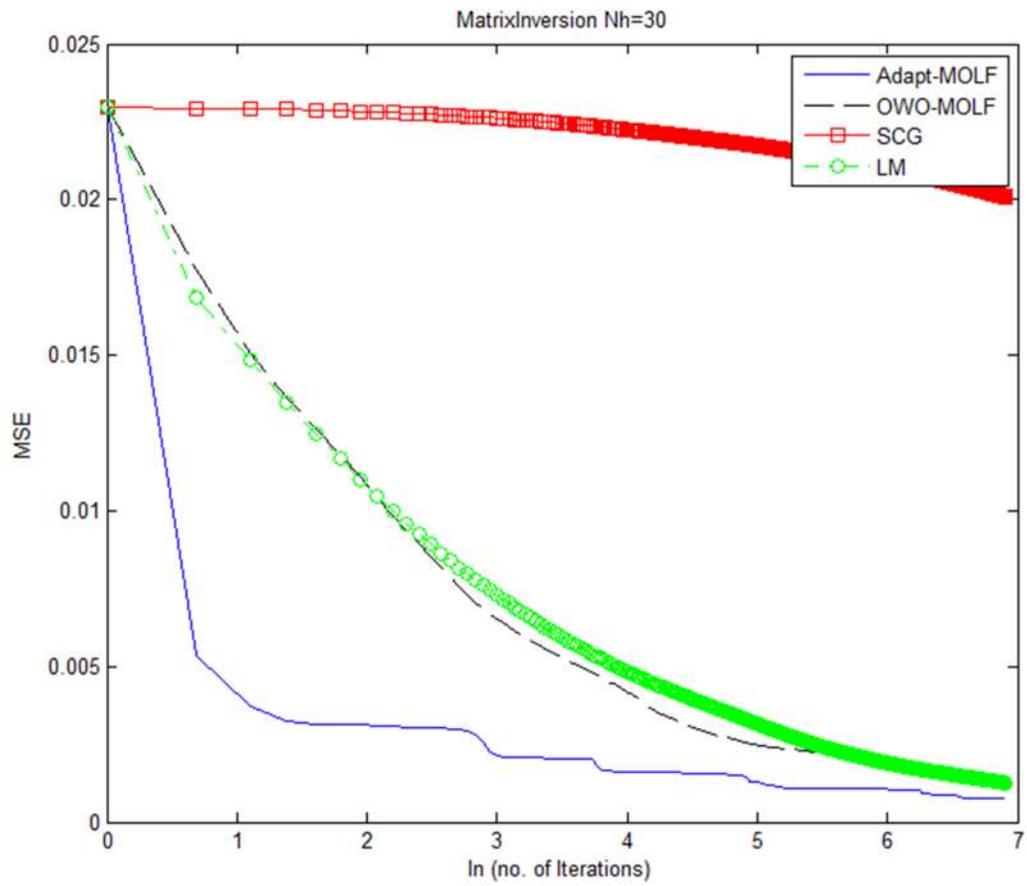

Figure 6.7 matrix inversion data set: average error vs. number of iterations



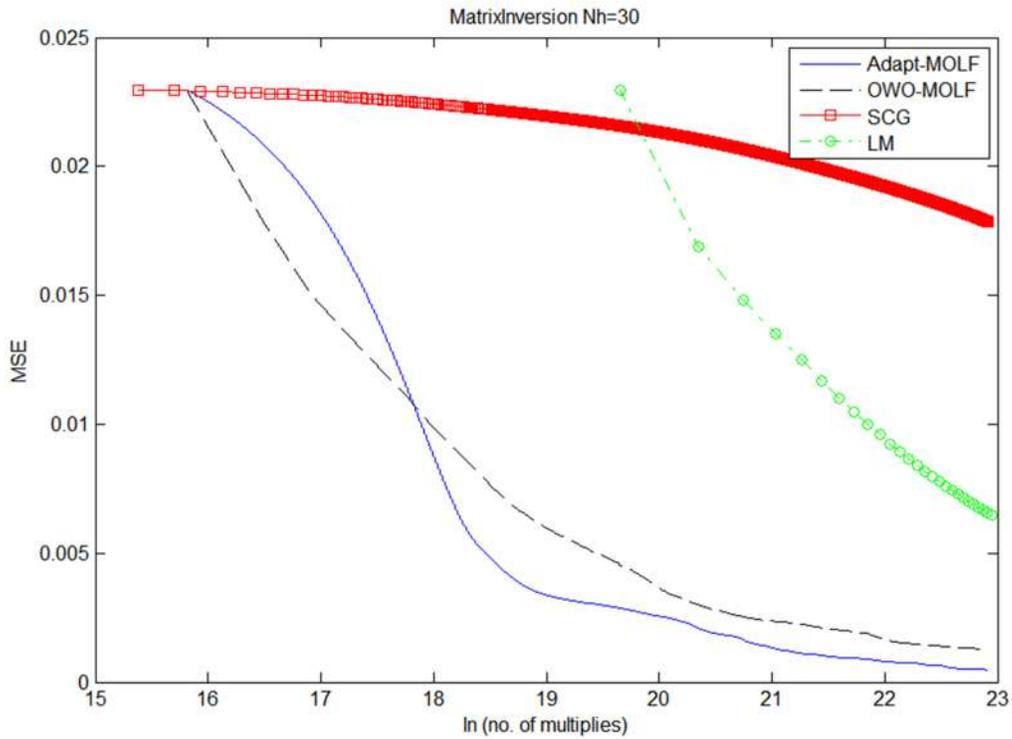

Figure 6.8 matrix inversion data set: average error vs. number of multiplies

For the matrix inversion data file [76], the MLP is trained with 30 hidden units. In Figure 6.7, the average mean square error (MSE) from 10-fold training is plotted versus the number of iterations for each algorithm (shown on a $\log_e$ scale). In Figure 6.8, the average training MSE from 10-fold training is plotted versus the required number of multiplies (shown on a $\log_e$ scale). For this dataset the proposed algorithm is dominating the other three algorithms.



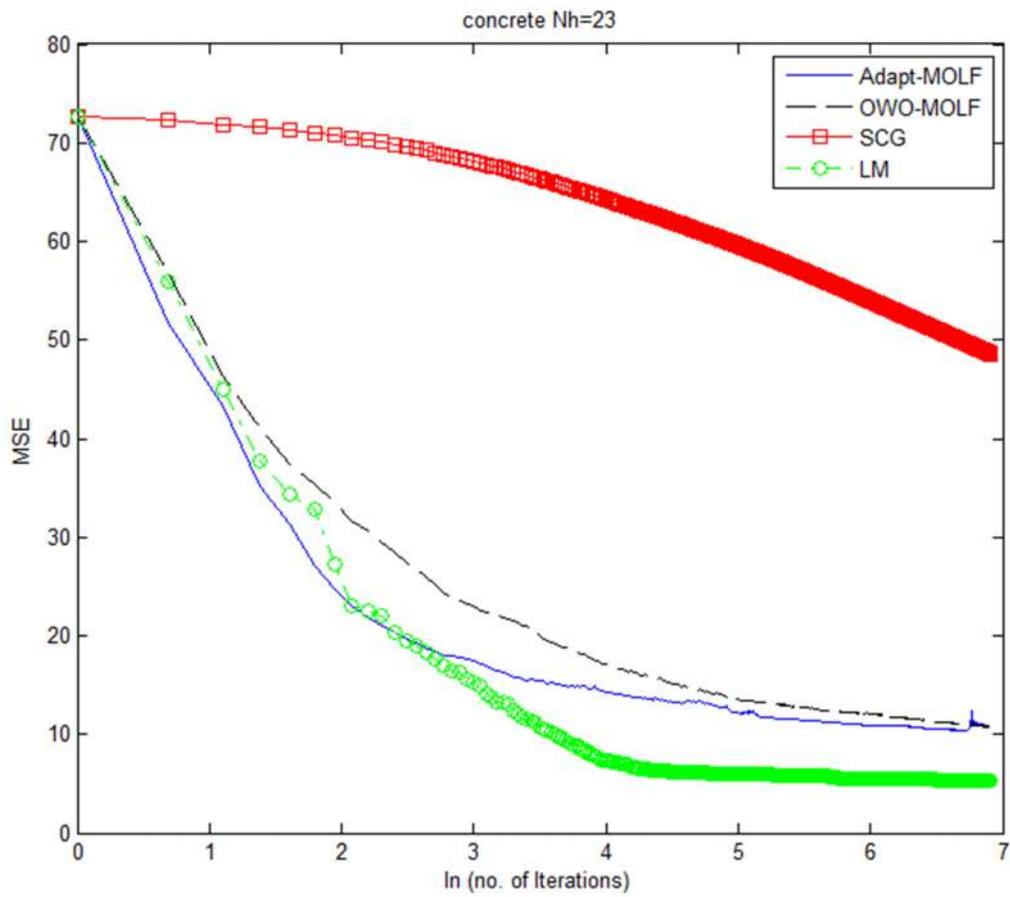

Figure 6.9 Concrete data set: average error vs. number of iterations



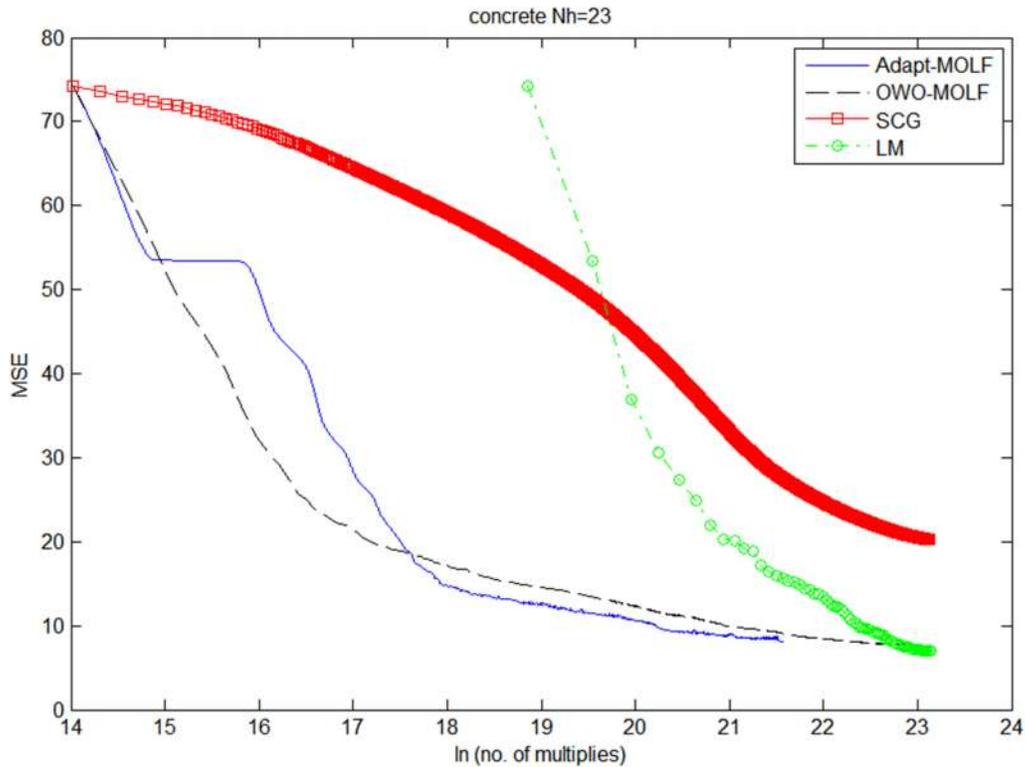

Figure 6.10 Concrete data set: average error vs. number of multiplies

For the concrete data file [77], the MLP is trained with 23 hidden units. In Figure 6.9, the average mean square error (MSE) for training from 10-fold training is plotted versus the number of iterations for each algorithm (shown on a $\log_e$ scale). In Figure 6.10, the average training MSE from 10-fold training is plotted versus the required number of multiplies (shown on a $\log_e$ scale). In this data set the proposed algorithm is performing better than OWO-MOLF in terms of iterations. In terms of multiples the performance of adaptive MOLF is very close to that of OWO-MOLF.

## 6.2 K-fold Validation and Testing

The k-fold validation procedure is used to obtain the average training and testing errors. In k-fold validation, given a data set, it is randomly split into k non-overlapping



parts of equal size, of which (k-2) parts are used for training, one part for validation and the remaining one part for testing. In this technique the training is stopped when we get a satisfactory validation error and the resulting network will be tested on the testing data to obtain the test error. This procedure is repeated k times in order to obtain average training and testing errors. For the simulations the k value is chosen as 10. Table 6.2 given below, compares the average training and testing errors of the adaptive MOLF algorithm with other algorithms for different data files.

Table 6.2 Average 10-fold training and testing errors

| Data Set | | Adaptive MOLF | OWO-MOLF | SCG | LM |
|---|---|---|---|---|---|
| Twod.tra | $E_{trn}$ | 0.0888 | 0.1554 | 1.0985 | 0.2038 |
| | $E_{tst}$ | 0.1172 | 0.1731 | 1.0945 | 0.2205 |
| Single2.tra | $E_{trn}$ | 0.0042 | 0.0151 | 3.5719 | 0.0083 |
| | $E_{tst}$ | 0.2319 | 0.1689 | 3.6418 | 0.0178 |
| Mattrn.tra | $E_{trn}$ | 0.0011 | 0.0027 | 4.2400 | 0.0022 |
| | $E_{tst}$ | 0.0013 | 0.0032 | 4.3359 | 0.0027 |
| Oh7.tra | $E_{trn}$ | 1.2507 | 1.3205 | 4.1500 | 1.1602 |
| | $E_{tst}$ | 1.4738 | 1.4875 | 4.1991 | 1.4373 |
| Housing | $E_{trn}$ | 2.7274 | 3.1318 | 5.9245 | 0.9543 |
| | $E_{tst}$ | 19.0627 | 24.6750 | 80.8080 | 133.990 |

From the plots and the Table presented above, it can be inferred that adaptive MOLF performs better than OWO-MOLF, LM and SCG algorithms both in terms of iteration and multiplies in most of the data sets.



Chapter 7

Conclusion and Future Work

The question of the number of learning factors actually needed by a training algorithm is addressed by introducing an algorithm that can adaptively change the number of learning factors computed in order to produce better error decrease per multiply. The performance of adaptive MOLF is superior to the OWO-MOLF algorithm in terms of error decrease per iteration and often in terms of error decrease per multiply. The proposed algorithm is found to interpolate between the OWO-MOLF and OWO-Newton algorithms. In some cases little is gained by increasing the number of learning factors beyond $N_h$.

In this thesis the concept of adaptive multiple optimal learning factors is applied only for input weights, this can be extended to output and bypass weights as well. There is a very good possibility that adaptive MOLF and similar algorithms will be successfully utilized in training auto encoders and for constructing deep neural networks.



Appendix A

Description of Data Sets Used For Training and Validation



*TWOD.TRA: (8 Inputs, 7 Outputs, 1768 Training Patterns, 244 KB)*

This training file is used in the task of inverting the surface scattering parameters from an inhomogeneous layer above a homogeneous half space, where both interfaces are randomly rough. The parameters to be inverted are the effective permittivity of the surface, the normalized rms height, the normalized surface correlation length, the optical depth, and single scattering albedo of an inhomogeneous irregular layer above a homogeneous half space from back scattering measurements.

The training data file contains 1768 patterns. The inputs consist of eight theoretical values of back scattering coefficient parameters at V and H polarization and four incident angles. The outputs were the corresponding values of permittivity, upper surface height, lower surface height, normalized upper surface correlation length, normalized lower surface correlation length, optical depth and single scattering albedo which had a joint uniform pdf.

*SINGLE2.TRA: (16 Inputs, 3 Outputs, 10,000 Training Patterns, 1.6MB)*

This training data file consists of 16 inputs and 3 outputs and represents the training set for inversion of surface permittivity, the normalized surface rms roughness, and the surface correlation length found in back scattering models from randomly rough dielectric surfaces. The first 16 inputs represent the simulated back scattering coefficient measured at 10, 30, 50 and 70 degrees at both vertical and horizontal polarization. The remaining 8 are various combinations of ratios of the original eight values. These ratios correspond to those used in several empirical retrieval algorithms.

*OH7.TRA: (20 Inputs, 3 Outputs, 15,000 Training Patterns, 3.1 MB)*

This data set is given in Oh, Y., K. Sarabandi, and F.T. Ulaby, "An Empirical Model and an Inversion Technique for Radar Scattering from Bare Soil Surfaces," in IEEE Trans. on Geoscience and Remote Sensing, pp. 370-381, 1992. The training set



contains VV and HH polarization at L 30, 40 deg, C 10, 30, 40, 50, 60 deg, and X 30, 40, 50 deg along with the corresponding unknowns rms surface height, surface correlation length, and volumetric soil moisture content in g / cubic cm.

*MAT.TRN: (4 Inputs, 4 Outputs, 2000 Training Patterns, 644KB)*

This training file provides the data set for inversion of random two-by-two matrices. Each pattern consists of 4 input features and 4 output features. The input features, which are uniformly distributed between 0 and 1, represent a matrix and the four output features are elements of the corresponding inverse matrix. The determinants of the input matrices are constrained to be between .3 and 2.

*Concrete Compressive Strength Data Set: (8 Inputs, 1 Output, 1030 Training Patterns)*

This data said is obtained from machine learning data set library provided by school of Information and Computer Science at University of California Irvine. This data set contains attributes like quantity of cement, blast furnace slag, fly ash, water, super plasticizer, coarse aggregate, fine aggregate, age and concrete compressive strength